\documentclass[letterpaper, 10pt, conference]{ieeeconf}
\IEEEoverridecommandlockouts
\usepackage{amsfonts,amsmath,amssymb}
\usepackage{enumerate}
\usepackage{graphicx}
\usepackage[hyphens]{url}
\usepackage{hyperref}
\usepackage{algorithm}
\usepackage{algorithmic}
\usepackage{cite}
\usepackage{gensymb}

\newtheorem{remark}{Remark}

\newcommand{\R}{\mathbb{R}}
\newcommand{\E}{\mathbb{E}}

\newcommand{\bK}{\mathbf{K}}
\newcommand{\bQ}{\mathbf{Q}}
\newcommand{\bS}{\mathbf{S}}
\newcommand{\btheta}{\boldsymbol{\theta}}
\newcommand{\bphi}{\boldsymbol{\phi}}
\newcommand{\beps}{\boldsymbol{\epsilon}}

\newcommand{\bSigma}{\boldsymbol{\Sigma}}
\newcommand{\cD}{\mathcal{D}}
\newcommand{\cN}{\mathcal{N}}
\newcommand{\cR}{\mathcal{R}}
\newcommand{\cJ}{\mathcal{J}}
\newcommand{\cZ}{\mathcal{Z}}
\newcommand{\cL}{\mathcal{L}}

\title{\LARGE \bf
Function-Space Priors for Bayesian Neural ODEs with Application to Vessel Trajectory Prediction
}

\author{Jaeyeong Lee$^{a}$, Wonmo Koo$^{a}$, and Heeyoung Kim$^{a,\dagger}$%
  \thanks{$^{a}$Department of Industrial and Systems Engineering,
    Korea Advanced Institute of Science and Technology (KAIST),
    Daejeon, Republic of Korea.}%
  \thanks{$^\dagger$Corresponding author:
    \href{mailto:heeyoungkim@kaist.ac.kr}{heeyoungkim@kaist.ac.kr}}%
}

\begin{document}

\maketitle
\thispagestyle{empty}
\pagestyle{empty}

\begin{abstract}
Vessel trajectory prediction from Automatic Identification System (AIS) data is essential for maritime situational awareness, yet it remains challenging due to irregular sampling, missing reports, and complex dynamics. Beyond accurate point forecasts, maritime applications also demand well-calibrated uncertainty estimates for reliable decision-making. Bayesian Neural Ordinary Differential Equations (ODEs) offer a principled framework for continuous-time trajectory modeling with uncertainty quantification by placing a prior over the neural vector field parameters. However, the commonly used isotropic Gaussian weight prior fails to encode informative structural properties of vessel dynamics, such as smoothness and locality. Existing function-space Bayesian neural network methods address this limitation for static mappings, but do not transfer directly to Neural ODEs, where the primary quantity of interest is the trajectory rather than the vector field itself. In principle, one could place a Gaussian process (GP) prior directly over ODE solutions, but this requires propagating distributions through a nonlinear ODE solver, which is analytically intractable. To address this challenge, we adopt a practical approach that imposes a GP-kernel-based prior directly on the vector field evaluated at a finite set of measurement points. 
Specifically, we augment the standard weight-space variational objective with a kernel-based regularizer that penalizes deviations of the vector field from the structure implied by a GP prior. To handle long and irregular AIS trajectories, we further combine this function-space regularization with probabilistic multiple shooting, which decouples inference across temporal segments while maintaining global consistency. We evaluate the proposed method on real-world AIS datasets and compare it against weight-space Bayesian Neural ODEs and GP-based ODE baselines. Experimental results demonstrate improvements in both predictive accuracy and uncertainty calibration for vessel trajectory prediction.
\end{abstract}

\section{Introduction}
\label{sec:intro}
The analysis of spatio-temporal data has attracted significant attention across many fields because of its importance for understanding and forecasting dynamic phenomena over geographic space and time \cite{koo2024deep,chung2019crime}. 
In particular, Automatic Identification System (AIS) data provides time-stamped vessel positions and kinematic states that enable large-scale maritime traffic monitoring \cite{tu2017exploiting, yang2019big,park2020maritime}. Accurate trajectory prediction from AIS underpins safety-critical tasks, including collision-risk assessment, traffic management, route planning, and anomaly detection \cite{kim2021locally,lee2023semi,kim2023contextual}. Despite substantial progress in sequence-based trajectory forecasting, AIS-based prediction remains difficult in practice: reports are often irregularly sampled, may contain gaps due to reception limitations, and exhibit highly nonlinear movements driven by navigation rules, weather conditions, and complex local interactions \cite{soh2018application,choy2016looking,lee2013dependence}. 
Maritime decision-making therefore requires not only accurate predictions but also reliable uncertainty estimates \cite{yoon2024uncertainty,yoon2026uncertainty}.

Continuous-time dynamical models, including Neural Ordinary Differential Equations (ODEs)~\cite{chen2018neural}, provide a natural framework for irregularly sampled trajectories by defining continuous dynamics between observations rather than relying on fixed time steps \cite{oh2025neural,oh2025dualdynamics}. Bayesian extensions place a prior over the vector field parameters and perform posterior inference, enabling principled (epistemic) uncertainty quantification. Recent work has combined Bayesian Neural ODEs with multiple shooting techniques to improve training stability on long sequences~\cite{iakovlev2022latent}, while Gaussian process (GP) ODEs (GPODEs)~\cite{hegde2022variational} directly model the vector field as a GP to obtain function-space posteriors.

A key design choice in Bayesian Neural ODEs is where to place the prior. The standard approach uses a weight-space prior, 
which is computationally convenient but does not directly encode interpretable properties of the vector field such as smoothness or spatial locality. Function-space Bayesian neural network (fBNN) methods~\cite{sun2019functional,rudner2022tractable,cinquin2025well} address this for static input--output mappings by placing priors directly over function outputs. However, extending this idea to Neural ODES presents a fundamental challenge: the observable quantity is the trajectory rather than the vector field  $f_{\btheta}$ itself.

Placing a GP prior over trajectories would require propagating uncertainty through an ODE solver, i.e., computing the distribution of ODE solutions induced by uncertainty in $\btheta$, which is intractable in closed form. Moreover, the linearization trick \cite{rudner2022tractable} to obtain a tractable function-space KL divergence relies on a first-order Taylor expansion of network outputs. Applying this idea to ODE solutions would require differentiating through the solver, which can become numerically unstable for long integration intervals. GPODEs~\cite{hegde2022variational} avoid this by modeling $f_{\btheta}$ directly as a GP, achieving function-space posteriors naturally. However, using standard GP with basic kernels can limit expressive capacity and leads to cubic computational scaling in the number of inducing points. 
This raises a practical question: \emph{Can we impose interpretable structural bias on the vector field of a Bayesian Neural ODE---without requiring a tractable trajectory-level prior---and does doing so improve uncertainty-aware vessel trajectory prediction under realistic AIS conditions?}

\begin{figure*}[h]
  \centering
  \includegraphics[width=0.53\textwidth]{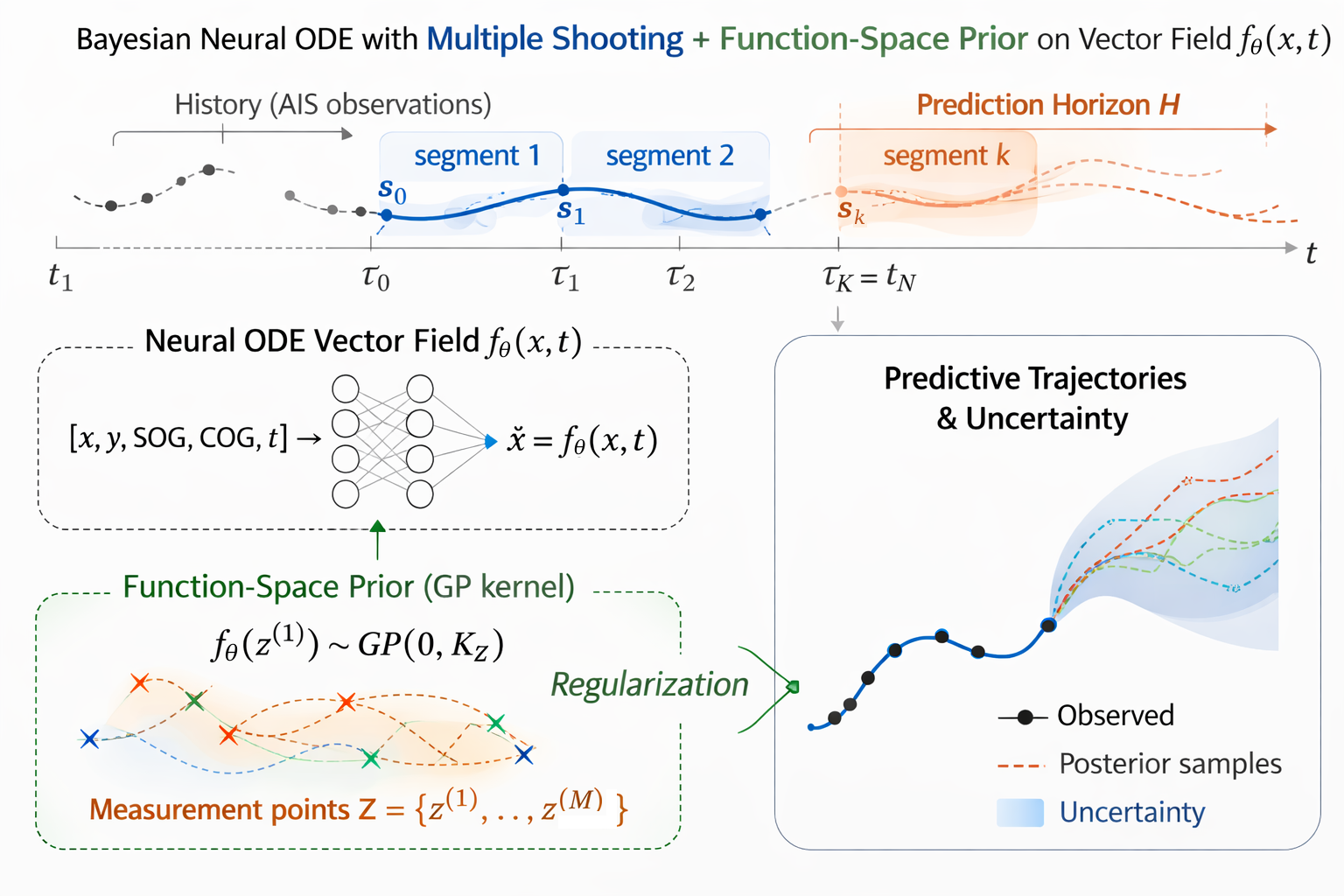}
  \caption{Model overview: probabilistic multiple shooting with a function-space (GP-kernel) prior on the neural vector field evaluated at a finite measurement set $\mathcal{Z}$.}
  \label{fig:overview}
\end{figure*}

In this paper, we explore function-space priors for Neural ODE-based vessel trajectory prediction via a practical approach that augments the standard weight-space variational objective with a kernel-based function-space regularizer evaluated at measurement points in the state space. To enable scalable inference, we combine this regularization with probabilistic multiple shooting. Our contributions are as follows:
\begin{enumerate}[(i)]
  \item \textbf{Function-space regularization for Bayesian Neural ODEs:}
    We incorporate a GP-kernel-based prior on the vector field into the variational objective, enabling interpretable structural bias on the learned dynamics while retaining the flexibility of neural network parameterizations.
  \item \textbf{Empirical comparison of prior strategies:} We systematically compare function-space priors against weight-space Bayesian Neural ODEs~\cite{iakovlev2022latent} and GPODEs~\cite{hegde2022variational} on real-world AIS data.
  \item \textbf{Analysis of measurement point design:} We investigate practical strategies for selecting measurement points and assess their effect on prediction quality.
\end{enumerate}
Fig.~\ref{fig:overview} provides an overview of our Bayesian Neural ODE framework for AIS trajectory prediction.
Our approach combines probabilistic multiple shooting for stable learning on long, irregular trajectories with a GP-kernel-based function-space regularizer that imposes structural bias directly on the neural vector field at a finite set of measurement points.

\section{Methodology}
\label{sec:method}

\subsection{Problem Setup}
\label{sec:problem}

We consider vessel trajectories observed via AIS as irregularly sampled multivariate time series. For each trajectory, we observe
\begin{equation}
  \cD = \{(t_i, \mathbf{y}_i)\}_{i=1}^{N}, \quad 0 \le t_1 < \cdots < t_N,
\end{equation}
where $t_i \in \R_{\ge 0}$ are observation times and $\mathbf{y}_i \in \R^{d}$ are observations:
\begin{equation}
  \mathbf{y}_i = [x_i,\, y_i,\, \mathrm{SOG}_i,\, \sin(\mathrm{COG}_i),\, \cos(\mathrm{COG}_i)]^\top \in \mathbb{R}^5,
\end{equation}
where $(x_i, y_i)$ are position coordinates in a local tangent plane (ENU), SOG is speed over ground (m/s), and COG is course over ground encoded via sine-cosine to avoid discontinuity at $0\degree / 360\degree$. Our goal is to predict the future trajectory $\{\mathbf{y}(t)\}_{t \in (t_N,\, t_N+H]}$ and quantify predictive uncertainty under irregular sampling and missingness.

\subsection{Latent ODE for Vessel Dynamics}
\label{sec:latentode}

We model the latent vessel state $\mathbf{x}(t)\in\R^{d}$ as a continuous-time dynamical system:
\begin{equation}
  \frac{d\mathbf{x}(t)}{dt} = f_{\btheta}(\mathbf{x}(t), t),
  \label{eq:ode}
\end{equation}
where $f_{\btheta}: \R^d \times \R_{\ge 0} \to \R^d$ is a neural vector field parameterized by $\btheta$. Given an initial state $\mathbf{x}(t_a)$, the state at time $t_b > t_a$ is obtained by ODE integration:
\begin{equation}
  \mathbf{x}(t_b) = \Phi_{t_a \to t_b}\!\big(\mathbf{x}(t_a); \btheta\big),
\end{equation}
where $\Phi$ denotes the flow map induced by an ODE solver.

To quantify epistemic uncertainty, we place a prior over the vector-field parameters $\btheta \sim p(\btheta)$ and perform approximate Bayesian inference to obtain a variational posterior $q_{\bphi}(\btheta) \approx p(\btheta \mid \cD)$.

\textbf{Observation model.}
We connect latent states to AIS measurements via $\mathbf{y}_i = g(\mathbf{x}(t_i)) + \beps_i$,
with a Gaussian likelihood:
\begin{equation}
  p(\mathbf{y}_i \mid \mathbf{x}(t_i))
  = \cN\!\big(\mathbf{y}_i;\, g(\mathbf{x}(t_i)),\, \bSigma\big),
  \label{eq:likelihood}
\end{equation}
with diagonal $\bSigma$ (learned or fixed) and $g(\cdot)$ the identity in
our experiments.

\subsection{Probabilistic Multiple Shooting}
\label{sec:shooting}

Direct Bayesian learning of Eq.~(\ref{eq:ode}) over long, irregular trajectories is unstable because gradients and solver errors accumulate over extended integration intervals. We adopt a probabilistic multiple shooting formulation.

We partition $[t_1, t_N]$ into $K$ segments with boundary times $t_1 = \tau_0 < \tau_1 < \cdots < \tau_K = t_N$, and introduce latent boundary states $\mathbf{s}_k \approx \mathbf{x}(\tau_k)$. Within each
segment $k$, the latent trajectory is:
\begin{equation}
  \mathbf{x}_k(t) = \Phi_{\tau_k \to t}\!\big(\mathbf{s}_k;\, \btheta\big),
  \quad t \in [\tau_k, \tau_{k+1}].
  \label{eq:segment_flow}
\end{equation}

\textbf{Probabilistic continuity constraint.}
To preserve global temporal consistency, we impose a soft continuity constraint:
\begin{equation}
  p(\mathbf{s}_{k+1} \mid \mathbf{s}_{k}, \btheta)
  = \cN\!\Big(\mathbf{s}_{k+1};\,
    \Phi_{\tau_k \to \tau_{k+1}}(\mathbf{s}_k;\btheta),\,
    \bQ_k\Big),
  \label{eq:shooting_prior}
\end{equation}
where $\bQ_k$ controls the tolerance for mismatch between consecutive segments. When $\bQ_k \to \mathbf{0}$, this reduces to deterministic shooting; larger $\bQ_k$ provides robustness to solver error and model mismatch.

\subsection{Function-Space Prior on the Vector Field}
\label{sec:fspace_prior}

A standard weight-space prior ($p(\btheta) = \cN(\mathbf{0}, \sigma^2\mathbf{I})$) regularizes the parameters but does not directly encode structural properties of the vector field. 
To inject interpretable inductive bias on the dynamics, we introduce a function-space prior term evaluated at a set of measurement points.

\textbf{Measurement set.}
Let $\cZ = \{\mathbf{z}^{(m)}\}_{m=1}^{M}$ be a fixed set of measurement
points in $\R^d$. We collect vector field evaluations into
\begin{equation}
  \mathbf{F}_{\cZ}(\btheta)
  \triangleq
  \big[f_{\btheta}(\mathbf{z}^{(1)}), \ldots, f_{\btheta}(\mathbf{z}^{(M)})\big]^\top
  \in \R^{M \times d}.
  \label{eq:fz}
\end{equation}

\textbf{Function-space prior.}
We impose a GP prior on $f$ with a separable kernel structure:
\begin{equation}
  \mathrm{vec}\!\big(\mathbf{F}_{\cZ}(\btheta)\big)
  \sim \cN\!\big(\mathbf{0},\, \bK_{\cZ} \otimes \mathbf{I}_d\big),
  \label{eq:fs_prior}
\end{equation}
where $\bK_{\cZ} \in \R^{M \times M}$ has entries $[\bK_{\cZ}]_{ij} = k(\mathbf{z}^{(i)}, \mathbf{z}^{(j)})$ for a chosen kernel (e.g.\ squared exponential or Mat\'{e}rn), and $\otimes$ denotes the Kronecker product. The separable structure reduces the kernel inverse from $\mathcal{O}(M^3 d^3)$ to $\mathcal{O}(M^3)$.

The corresponding function-space regularizer decomposes across output
dimensions:
\begin{align}
  \cR_{\mathrm{FS}}(\btheta)
  &\triangleq
  -\log \cN\!\big(\mathrm{vec}(\mathbf{F}_{\cZ}(\btheta));\,
    \mathbf{0},\, \bK_{\cZ} \otimes \mathbf{I}_d\big) \nonumber\\
  &= \frac{1}{2} \sum_{j=1}^{d} \mathbf{f}_j^\top \bK_{\cZ}^{-1} \mathbf{f}_j
  + \frac{d}{2}\log|\bK_{\cZ}|
  + \mathrm{const.},
  \label{eq:rfs}
\end{align}
where $\mathbf{f}_j = [f_{\btheta}^{(j)}(\mathbf{z}^{(1)}), \ldots, f_{\btheta}^{(j)}(\mathbf{z}^{(M)})]^\top \in \R^M$ denotes the $j$-th output dimension of the vector field at $\cZ$. The gradient $\nabla_{\mathbf{f}_j}\cR_{\mathrm{FS}} = \bK_{\cZ}^{-1}\mathbf{f}_j$ is backpropagated through the network via the chain rule.

\textbf{Choice of measurement points.}
The set $\cZ$ should represent the state-space region visited by typical
vessel trajectories. We consider two strategies:
\begin{enumerate}[(i)]
  \item \textbf{$k$-means on trajectories:} Apply $k$-means clustering to the training trajectory states to obtain $M$ representative centroids.
  \item \textbf{Maneuver-weighted sampling:} Oversample states near turning or deceleration events, where smoothness priors are most informative.
\end{enumerate}
In all cases, $\cZ$ is fixed after construction and does not depend on
$\btheta$ during training.

\textbf{Relation to function-space BNN methods.}
Recent function-space BNN methods~\cite{rudner2022tractable,cinquin2025well} achieve tractable function-space inference by placing priors directly over network \emph{outputs} and computing a regularized KL divergence in the output space. In the Neural ODE setting, however, the relevant output is the ODE \emph{solution} $\mathbf{x}(t) = \Phi_{t_0 \to t} (\mathbf{x}_0;\btheta)$ rather than a direct forward pass of $f_{\btheta}$.  Propagating a GP prior through the flow map $\Phi$ lacks a closed form, and the linearization trick of fSVI~\cite{rudner2022tractable} requires a first-order Taylor approximation of the ODE solution with respect to $\btheta$, which is numerically unstable over long integration intervals. We therefore impose the GP prior directly on the \emph{vector field} $f_{\btheta}$ at a finite measurement set $\cZ$, rather than on trajectories. This is a strictly weaker condition than a trajectory-level GP prior--- two vector fields consistent with the same GP at $\cZ$ may produce different trajectories---but it provides a tractable and interpretable inductive bias on the learned dynamics. The resulting objective is formally justified under the generalized variational inference framework of \cite{bissiri2016general} (see Remark~\ref{rem:gvi}).

\textbf{Complementary roles of weight-space and function-space regularization.}
The weight-space KL term provides standard parameter-level regularization, whereas $\cR_{\mathrm{FS}}(\btheta)$ directly injects structural inductive bias (e.g.\ smoothness encoded by the kernel) onto the vector field in function space. The two terms operate in complementary spaces and serve distinct roles.

\subsection{Regularized Variational Objective}
\label{sec:objective}

We perform variational inference with an approximate posterior:
\begin{equation}
  q_{\bphi}(\btheta, \bS)
  = q_{\bphi}(\btheta) \prod_{k=0}^{K} q_{\bphi}(\mathbf{s}_k),
  \label{eq:q_factor}
\end{equation}
where $q_{\bphi}(\btheta)$ and $q_{\bphi}(\mathbf{s}_k)$ are diagonal
Gaussians, and $\bS = \{\mathbf{s}_k\}_{k=0}^{K}$.

We maximize the following regularized variational objective:
\begin{align}
  \cJ(\bphi)
  &= \cL_{\mathrm{shoot}}(\bphi)
    - \lambda_{\mathrm{FS}}\,
      \E_{q_{\bphi}(\btheta)}\!\Big[\cR_{\mathrm{FS}}(\btheta)\Big],
  \label{eq:objective}
\end{align}
where $\cL_{\mathrm{shoot}}$ is the shooting ELBO:
\begin{align}
  \cL_{\mathrm{shoot}}
  &= \E_{q}\!\bigg[
      \sum_{i=1}^{N} \log p(\mathbf{y}_i \mid \mathbf{x}(t_i))
    + \sum_{k=0}^{K-1} \log p(\mathbf{s}_{k+1} \mid \mathbf{s}_{k}, \btheta)
    \bigg]
    \nonumber\\[2pt]
  &\quad
    + \sum_{k=1}^{K} H\!\big[q_{\bphi}(\mathbf{s}_k)\big]
    - \mathrm{KL}\!\big(q_{\bphi}(\mathbf{s}_0) \,\|\, p(\mathbf{s}_0)\big)
    \nonumber\\[2pt]
  &\quad
    - \mathrm{KL}\!\big(q_{\bphi}(\btheta) \,\|\, p(\btheta)\big),
  \label{eq:elbo_shoot}
\end{align}
where $H[q_{\bphi}(\mathbf{s}_k)] = -\E_{q}\!\big[\log q_{\bphi}(\mathbf{s}_k)\big]$ denotes the entropy of the shooting state posterior.
Setting $\lambda_{\mathrm{FS}} = 0$ recovers the standard weight-space Bayesian ODE with shooting. We optimize $\cJ(\bphi)$ using stochastic gradient ascent with the reparameterization trick and backpropagation through the ODE solver.

\begin{remark}
\label{rem:gvi}
Objective~(\ref{eq:objective}) can be interpreted as an evidence lower bound (ELBO) under a modified prior $\tilde{p}(\btheta) \propto p(\btheta) \exp\{-\lambda_{\mathrm{FS}}\,\cR_{\mathrm{FS}}(\btheta)\}$, connecting to the generalized Bayesian inference framework~\cite{bissiri2016general}.
This perspective avoids the infinite KL divergence issue identified by \cite{burt2020understanding}: the KL divergence between a parametric variational family and a non-degenerate GP prior is infinite, as the two induce mutually singular measures in function space. Rather than computing a function-space KL divergence directly, our approach introduces the regularizer $\cR_{\mathrm{FS}}$, which penalizes the vector field according to a GP log-density evaluated at $M$ measurement points, which is always finite and well-defined.
We treat $\lambda_{\mathrm{FS}}$ as a hyperparameter and select it based on validation performance.
\end{remark}

\subsection{Prediction and Uncertainty Quantification}
\label{sec:prediction}

Given observations up to $t_N$, we predict future states by sampling from
the variational posterior and integrating forward:
\begin{equation}
  \hat{\mathbf{x}}^{(s)}(t)
  = \Phi_{t_N \to t}\!\big(\mathbf{s}_K^{(s)};\, \btheta^{(s)}\big),
  \quad t > t_N,
  \label{eq:predictive_samples}
\end{equation}
with $\btheta^{(s)} \sim q_{\bphi}(\btheta)$ and
$\mathbf{s}_K^{(s)} \sim q_{\bphi}(\mathbf{s}_K)$.
The predictive distribution is approximated by a Monte Carlo mixture:
\begin{equation}
  p(\mathbf{y}(t)\mid\cD)
  \approx
  \frac{1}{S}\sum_{s=1}^{S}
  \cN\!\big(\mathbf{y}(t);\, g(\hat{\mathbf{x}}^{(s)}(t)),\, \bSigma\big).
  \label{eq:predictive_mixture}
\end{equation}
The spread of predictive samples captures epistemic uncertainty, while $\bSigma$ accounts for aleatoric uncertainty.

\subsection{Computational Considerations}
\label{sec:compute}

Multiple shooting restricts ODE integration to short intervals, so overall
cost scales approximately linearly with the number of segments. The function-space regularizer~(\ref{eq:rfs}) introduces an additional cost of $\mathcal{O}(M^3 + Md)$ per gradient step. Since $M$ is typically small (e.g.\ $M = 20$--$50$), this overhead is negligible in practice. The kernel matrix $\bK_{\cZ}^{-1}$ is precomputed once and reused across training
iterations when $\cZ$ is fixed.

\section{Related Work}
\label{sec:related}

\subsection*{Neural ODEs and Bayesian Extensions}

Neural ODEs~\cite{chen2018neural} parameterize continuous-time dynamics with neural networks and optimize via adjoint sensitivity, enabling flexible modeling of irregularly sampled time series. Rubanova et al.~\cite{rubanova2019latent} extended this to the latent ODE framework, encoding observations into a latent initial state and decoding through an ODE solver. However, standard Neural ODEs provide only point estimates and no principled uncertainty quantification.

Bayesian extensions address this limitation by placing priors over the vector field parameters. Dandekar et al.~\cite{dandekar2020bayesian} applied various Bayesian inference methods such as Hamiltonian Monte Carlo to Neural ODE weights, but this scales poorly to large models. Variational approaches offer a practical alternative: Iakovlev et al.~\cite{iakovlev2022latent} introduced probabilistic multiple shooting for Bayesian latent ODEs, splitting long integration intervals into short segments with learnable boundary states to mitigate gradient instability. Hegde et al.~\cite{hegde2022variational} proposed GPODE, which models the vector field directly as a GP, obtaining function-space posteriors naturally through decoupled functional sampling and sparse inducing variables. While GPODE provides well-calibrated uncertainty, using standard GP with basic kernel limits their capacity, and it scales cubically in the number of inducing points and poorly with state dimensionality.

Our work builds on this line by combining the flexibility of neural network vector fields with the structured uncertainty of Bayesian inference, augmented by function-space priors that encode interpretable properties of the dynamics.

\subsection*{Function-Space Bayesian Neural Networks}

A key limitation of weight-space priors such as $p(\btheta) = \cN(\mathbf{0}, \sigma^2\mathbf{I})$ is that they do not directly encode functional properties like smoothness or locality. Sun et al.~\cite{sun2019functional} proposed functional variational BNNs (fBNNs) that place priors directly over network outputs, but the resulting KL divergence is intractable for networks of different widths.
Burt et al.~\cite{burt2020understanding} formalized this obstacle, showing that the function-space KL divergence between a parametric variational posterior and a non-degenerate GP prior is infinite, as the two induce mutually singular measures on the function space.

Rudner et al.~\cite{rudner2022tractable} circumvented this via fSVI, which constrains the prior and posterior to share the same network architecture and uses a first-order linearization to obtain a tractable Gaussian approximation in function space. Cinquin and Bamler~\cite{cinquin2025well} further generalized this with GFSVI, introducing a regularized KL objective that admits meaningful GP priors without requiring identical architectures, connecting to the generalized variational inference framework \cite{bissiri2016general}.

These methods target static input--output mappings. In the Neural ODE
setting, the relevant output is the trajectory $\mathbf{x}(t) = \mathbf{x}_0 + \int f_{\btheta}\,ds$ rather than the vector field $f_{\btheta}$ itself. Propagating a GP prior through the ODE solver to obtain a trajectory-level prior is intractable, and the linearization trick of fSVI becomes numerically unstable over long integration intervals. To our knowledge, this is the first empirical exploration of function-space priors in the Neural ODE setting. We adopt a practical approach: rather than placing a prior on trajectories, we regularize the vector field at a finite set of measurement points, obtaining a tractable and interpretable inductive bias on the learned dynamics.

\section{Experiments}
\label{sec:experiments}

\subsection{Dataset and Preprocessing}
\label{sec:data}

We evaluate our approach on real-world AIS data from the New York/New Jersey Harbor
region, covering January 1--7, 2024. Each AIS message contains a timestamp, latitude/longitude position, SOG, and COG. We retain messages within the bounding box
$[40.40^\circ, 40.80^\circ]\mathrm{N} \times [74.30^\circ, 73.70^\circ]\mathrm{W}$, yielding approximately 1.5\,M records from 590 unique vessels. 
We apply standard preprocessing steps:
(i)~removing duplicate timestamps per vessel,
(ii)~discarding messages with invalid positions or kinematic values, and
(iii)~segmenting each vessel's record into contiguous trajectories using a 30-minute gap threshold, retaining only segments with at least 20 observations and 15 minutes duration. This process results in 1,589 trajectories.

\begin{figure}[t]
  \centering
  \includegraphics[width=0.9\columnwidth]{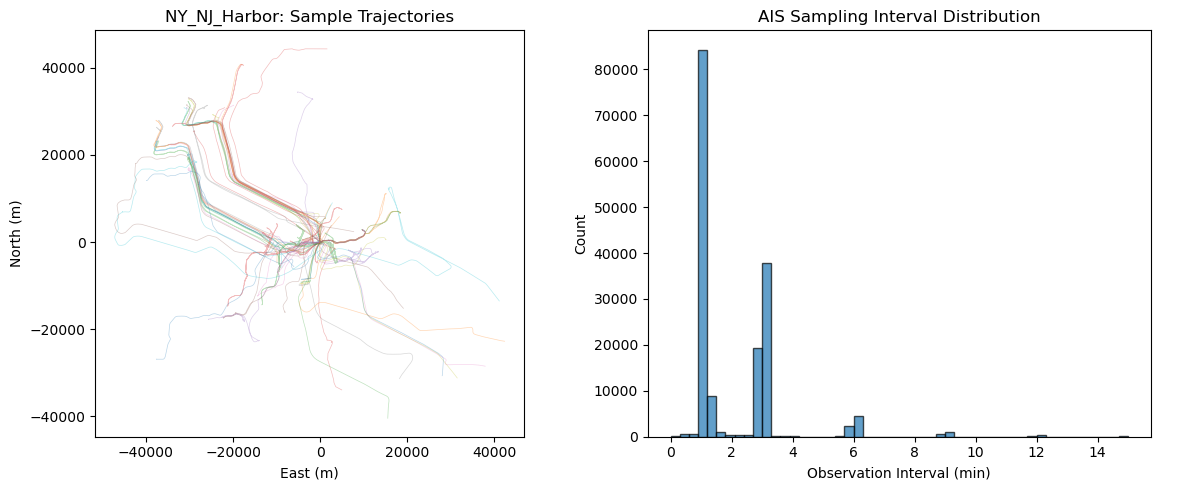}
  \caption{Left: Sample vessel trajectories in the NY/NJ Harbor region
    (ENU coordinates). Right: Distribution of AIS observation intervals,
    showing irregular sampling with intervals predominantly between 2 and 4 minutes, which is characteristic of AIS data.}
  \label{fig:data}
  \vspace{-0.1 in}
\end{figure}

Latitude/longitude coordinates are converted to a local East--North--Up
(ENU) tangent plane in meters, using each trajectory's first observation as the origin. SOG is converted from knots to m/s, and COG is encoded as $[\sin(\mathrm{COG}), \cos(\mathrm{COG})]$ to avoid discontinuity at $0^\circ/360^\circ$, giving a five-dimensional state
$\mathbf{y} = [x, y, \mathrm{SOG}, \sin\mathrm{COG}, \cos\mathrm{COG}]^\top \in \R^5$. 
We split the data by vessel identity (70/10/20\% train/validation/test) to
prevent leakage, and construct sliding windows with 10--20 minutes of
history and a 10-minute prediction horizon.

\subsection{Experimental Protocol}
\label{sec:protocol}

Unlike the single-trajectory setting common in ODE learning benchmarks~\cite{hegde2022variational,chen2018neural}, AIS-based forecasting requires predicting diverse vessel trajectories that differ in speed, heading, and navigational intent. A single global vector field cannot capture vessel-specific dynamics from kinematic states alone. 
We therefore adopt a \emph{per-trajectory inference} protocol.
In a prior-construction phase (Phase~1), we use the training set to fix
quantities shared across trajectories: the measurement point set~$\cZ$
(via $k$-means or maneuver-weighted sampling) and kernel hyperparameters.
At test time (Phase~2), each trajectory is processed independently:
\begin{enumerate}
  \item A fresh model is initialized.
  \item The model is fit to the trajectory's history observations
    (${\sim}5$--$10$ points over 10--20 minutes) for a fixed number of
    optimization steps.
  \item The fitted model predicts the future 10-minute trajectory, with posterior samples providing uncertainty estimates.
\end{enumerate}
Because the history windows are short, we set $K=1$ for all methods and focus on the effect of the function-space regularizer; multiple shooting mainly serves as a scalable extension for longer histories. All methods follow the same protocol to ensure a fair comparison.

\subsection{Baselines and Ablations}
\label{sec:baselines}

We compare four Bayesian continuous-time models under the per-trajectory
protocol:
\begin{itemize}
  \item \textbf{GPODE}~\cite{hegde2022variational}: A Gaussian process vector field
    with decoupled functional sampling and inducing variables initialized
    via empirical gradients from the history observations.
  \item \textbf{WS-BayesNODE}: A Bayesian Neural ODE with a standard
    isotropic Gaussian weight-space prior
    $p(\btheta) = \cN(\mathbf{0}, \mathbf{I})$.
  \item \textbf{FS-BayesNODE ($k$-means)}: The proposed method with a
    GP-kernel function-space regularizer, where measurement points are
    selected by $k$-means clustering on training trajectory states.
  \item \textbf{FS-BayesNODE (maneuver)}: The proposed method with
    measurement points concentrated near turning and deceleration events.
\end{itemize}

The vector field for the Bayesian Neural ODE variants is parameterized as a single-hidden-layer MLP with 32 units. We use Adam optimization with early stopping on validation NLL.

\subsection{Evaluation Metrics}
\label{sec:metrics}

We report both point-prediction accuracy and probabilistic forecast quality:
\begin{itemize}
  \item \textbf{ADE / FDE}: Average and final displacement error (km)
    between the predictive mean position and ground truth.
  \item \textbf{NLL}: Negative log-likelihood under the predictive
    mixture~(\ref{eq:predictive_mixture}), evaluated on position
    coordinates.
  \item \textbf{CRPS}: Continuous ranked probability score for position
    (km), measuring both sharpness and calibration.
\end{itemize}
Metrics are computed in raw scale (km for position, m/s for SOG) after unnormalizing predictions.
We report the mean and standard deviation over 40 test trajectories, each evaluated with 30 posterior samples.

\subsection{Results}
\label{sec:results}

Table~\ref{tab:main_results} summarizes the main results.

\begin{table}[t]
\centering
\caption{Prediction performance on AIS trajectories
  (10-min horizon). Mean\,(std) over 40 test trajectories.
  Best in \textbf{bold}, second best \underline{underlined}.}
\label{tab:main_results}
\footnotesize
\setlength{\tabcolsep}{3pt}
\begin{tabular}{lcccc}
\hline
 & ADE\,(km)$\downarrow$ & FDE\,(km)$\downarrow$
 & NLL$\downarrow$ & CRPS\,(km)$\downarrow$ \\
\hline
GPODE
  & 4.44\,(3.49) & 5.67\,(3.82)
  & 29.46\,(6.88) & 4.57\,(1.17) \\
WS-BayesNODE
  & 3.03\,(1.80) & 4.47\,(2.75)
  & \textbf{18.90}\,(1.26) & 1.51\,(0.71) \\
FS-BayesNODE (km)
  & \underline{2.80}\,(1.98) & \underline{3.97}\,(2.60)
  & 19.31\,(3.31) & \underline{1.46}\,(0.87) \\
FS-BayesNODE (mn)
  & \textbf{2.66}\,(1.69) & \textbf{3.82}\,(2.22)
  & \underline{18.96}\,(1.67) & \textbf{1.41}\,(0.66) \\
\hline
\end{tabular}
\end{table}

\textbf{Point prediction.}
Both FS-BayesNODE variants substantially outperform GPODE in point prediction accuracy. FS-BayesNODE (maneuver) achieves the lowest ADE (2.66\,km) and FDE (3.82\,km), reducing displacement error by 40\% relative to GPODE (4.44\,km). The weight-space variant (WS-BayesNODE) also improves over GPODE, but the function-space regularizer provides additional gains by constraining the vector field structure.

\textbf{Probabilistic forecasting.}
All three BayesNODE variants achieve substantially better NLL
(${\sim}18.9$ vs.\ 29.46 for GPODE) and CRPS (1.41--1.51\,km vs.\ 4.57\,km). The improvement in CRPS by 69\% demonstrates that the Bayesian Neural ODE posterior is far more concentrated around the ground truth than the GP posterior, which suffers from over-dispersed uncertainty in the per-trajectory fitting regime.

\textbf{Measurement point selection.}
Maneuver-weighted measurement points outperform $k$-means in ADE, FDE, and CRPS, confirming that concentrating $\cZ$ near states where vessels change heading or speed provides more informative regularization in dynamically complex regions. The $k$-means variant still improves over the weight-space baseline in point prediction, suggesting that even a generic placement of measurement points provides useful inductive bias through the function-space regularizer.

\begin{figure}[t]
 \vspace{-0.1 in}
  \centering
  \includegraphics[width=0.9\columnwidth]{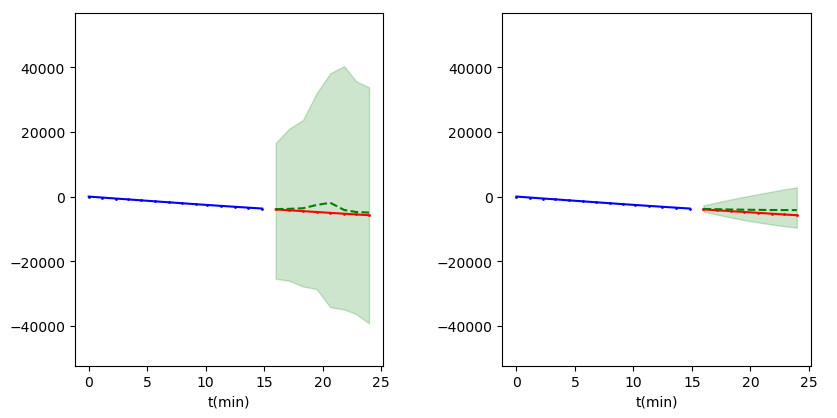}
   \vspace{-0.1 in}
  \caption{Predictive uncertainty comparison for a test trajectory.
    Left: GPODE. Right: FS-BayesNODE (maneuver).
    Blue: history; red: ground truth; green dashed: predictive mean;
    shaded: 90\% credible interval.
    The function-space prior produces uncertainty that grows smoothly with the prediction horizon, whereas GPODE uncertainty expands abruptly.}
  \label{fig:uncertainty}
  \vspace{-0.1 in}
\end{figure}

\textbf{Qualitative comparison.}
Figure~\ref{fig:uncertainty} illustrates the predictive uncertainty for a representative test trajectory. GPODE (left) produces uncertainty bands that expand rapidly over the prediction horizon, spanning tens of kilometers within a few minutes. In contrast, FS-BayesNODE (right) produces uncertainty that grows gradually, reflecting the natural increase in epistemic uncertainty with prediction distance. Both methods achieve comparable mean predictions (green dashed), but the function-space prior yields substantially sharper uncertainty.

\section{Conclusion}
\label{sec:conclusion}

In this paper, we explored function-space priors for Bayesian Neural ODE-based vessel trajectory prediction. By augmenting the standard weight-space variational objective with a GP-kernel-based regularizer evaluated at measurement points in state space, we inject interpretable structural bias---such as smoothness of the learned vector field---directly into the inference procedure. Experiments on real-world AIS data show that the proposed function-space prior improves both predictive accuracy and uncertainty calibration compared to weight-space Bayesian Neural ODEs and Gaussian process ODE baselines, with maneuver-weighted measurement points providing additional gains in maneuvering regimes and under sparse, irregular observations.

\section*{Acknowledgment}
This work was supported by the National Research Foundation of Korea (NRF) grant funded by the Korea government (MSIT) (2023R1A2C2005453, RS-2023-00218913).

\addtolength{\textheight}{-2cm}

\bibliographystyle{ieeetr}
\bibliography{bibliography}



\end{document}